# Definition drives design:
# Disability models and mechanisms of bias in AI technologies

Denis Newman-Griffis
Information School, University of Sheffield, Sheffield, UK
Department of Biomedical Informatics, University of Pittsburgh, Pittsburgh, PA, USA

Jessica Sage Rauchberg
Department of Communication Studies and Media Arts, McMaster University, Hamilton, Ontario, Canada

Rahaf Alharbi
School of Information, University of Michigan, Ann Arbor, MI, USA

Louise Hickman
Minderoo Centre for Technology and Democracy, University of Cambridge, Cambridge, UK

Harry Hochheiser
Department of Biomedical Informatics, University of Pittsburgh, Pittsburgh, PA, USA
Intelligent Systems Program, University of Pittsburgh, Pittsburgh, PA, USA

**Abstract**

The increasing deployment of artificial intelligence (AI) tools to inform decision making across diverse areas including healthcare, employment, social benefits, and government policy, presents a serious risk for disabled people, who have been shown to face bias in AI implementations. While there has been significant work on analysing and mitigating algorithmic bias, the broader mechanisms of how bias emerges in AI applications are not well understood, hampering efforts to address bias where it begins. In this article, we illustrate how bias in AI-assisted decision making can arise from a range of specific design decisions, each of which may seem self-contained and non-biasing when considered separately. These design decisions include basic problem formulation, the data chosen for analysis, the use the AI technology is put to, and operational design elements in addition to the core algorithmic design. We draw on three historical models of disability common to different decision-making settings to demonstrate how differences in the definition of disability can lead to highly distinct decisions on each of these aspects of design, leading in turn to AI technologies with a variety of biases and downstream effects. We further show that the potential harms arising from inappropriate definitions of disability in fundamental design stages are further amplified by a lack of transparency and disabled participation throughout the AI design process. Our analysis provides a framework for critically examining AI technologies in decision-making contexts and guiding the development of a design praxis for disability-related AI analytics. We put forth this article to provide key questions to facilitate disability-led design and participatory development to produce more fair and equitable AI technologies in disability-related contexts.

*Correspondence to:* Denis Newman-Griffis (d.r.newman-griffis@sheffield.ac.uk)

**Introduction**

Artificial intelligence (AI) technologies are increasingly being deployed to analyse information and support decision making in diverse areas, from healthcare and social services to employment, policy making, and urban planning. In many of these areas, disabled people[1] experience considerable injustice and ableism[2] in how they are treated, how they are perceived, and how they are included or excluded from decision-making and broader participation (Shakespeare et al, 2009; Ross and Taylor, 2017). Incorporating AI tools into these decision-making processes, when studies have shown that inappropriately-designed AI may already be biased against disabled people (Hutchinson et al, 2019; Kamikubo et al, 2022), thus presents serious risks of further exclusion and harm.

However, we argue that bias and ableism are neither inherent nor unavoidable products of using AI technologies in decision making. Nor are these outcomes a result of algorithmic engineering alone. Rather, bias and ableism arise as a result of specific design elements and decisions made throughout the process of developing, evaluating, and managing AI technologies in particular, situated contexts. In this article, we put forth a critical framework for identifying and investigating a key set of these design elements and decisions, as an important step in developing a broader methodology for critically interrogating AI technology design and use.

We analyse the first step of developing AI technologies, the process of *problem formulation*. Our framework breaks this process down into the component parts of defining an AI technology's scope, identifying the data it will use, defining the use it will be put to in context, and creating the operational definition the technology will fulfil. To illustrate how this framework can guide design and critical analysis, we analyse two realistic use cases for disability-related AI analytics: collaborative priority-setting and decision-making in healthcare, and evidence review for eligibility assessment in government benefits programs. Technology development for each of these use cases depends heavily on an operating definition of what disability is, and what is considered relevant information about it. We show the importance of acknowledging and directly interrogating these definitions—and the design decisions they inform—by designing speculative AI technologies for each of these use cases under three different conceptualisations of what "disability" means and how it may be represented or measured.

We draw on three historical conceptual models of disability that disabled people still commonly encounter today—the medical model, the social model, and the relational model—to demonstrate how AI development for the same goals and in the same contexts, but with different definitions of disability, produces highly distinct technologies with different biases and implications for their potential use as "black boxes" for data analytics. We show that technologies designed under different models may use the same data for different purposes, such as analysing medical records for information about documented support needs under the relational model, functional limitations and environmental barriers under the social model, or medical conditions and impairments under the medical model. Similarly, AI technologies may be designed to use entirely different sources of information, and produce entirely different types of insights, depending on the conceptualisation of disability used to guide development.

Our analysis demonstrates how individual choices in technology design have the aggregate effect of embodying specific ideologies in the resulting AI tools, and thus reflecting those same ideologies in the decisions that are informed by those AI tools. These design choices, and AI design processes more broadly, are often guided by unspoken or unconscious ideas and perceptions that are assumed to be self-evident or common sense, and thus left unquestioned. Our analysis illustrates the critical role of actively interrogating the assumptions that inform AI development and demonstrates how our framework can serve as a starting point for critical reflection in AI development and assessment.

The structure of our analysis draws on a multidisciplinary base of perspectives and methodologies and reflects interactions between a range of components. To provide this necessary context for our discussion, we first situate it with respect to our own positionalities as authors and the academic disciplines our discussion draws on and informs. We then highlight some of the observed mechanisms that support ableism in AI development and show how these illustrate the need to interrogate ideology in AI design. Motivated by this need, we introduce the components of our analysis, including our two use cases for AI development and the three conceptualisations of disability which we use to illustrate the application of our framework. With this base in place, we describe the components of our framework and the questions and design elements they address, and provide illustrations of how these components interact with our use cases and example conceptualisations of disability at each step. We conclude by discussing the implications and limitations of our current framework and prospective steps towards expanding and improving it as part of a broader design praxis for disability-related AI.

**Positionality and the multi-disciplinary dialogue of AI and disability**

To situate our work, we first describe our own positionality and the academic discourses our analysis aligns with. We are a multidisciplinary team of authors in North America and the United Kingdom, including authors trained in computer science and health informatics as well as critical data studies and disability studies. Our team includes authors who self-identify as disabled and authors who self-identify as non-disabled. We further represent multiple queer identities and racial identities (though a majority of authors identify as White), and we ground our analysis in an intersectional perspective. All authors have worked at the intersection of AI technologies and disability, including both designing disability-focused AI technologies and critical analysis of AI practices from a disability studies perspective. Although we draw largely on research and experience in the US and UK, our analysis of AI technologies as situated artefacts that are co-produced with their context of use has broader implications and can inform AI development priorities in Europe (European Commission, 2020), Africa (Nayebare, 2019; Gwagwa et al, 2020), and Asia (Gal, 2020; Younas, 2020) as well.

Our analysis of AI development from a sociotechnical perspective interacts with academic dialogue across multiple disciplines. Our definition and discussion of specific questions and challenges in the AI development process is grounded in the AI and machine learning research literature, including theory and methodology as well as situated applications. Several of the technical analyses we draw on focus on characterising, measuring, and mitigating bias and inequity in the area of AI methodology and modelling, such as demonstrating gender and/or racial bias in pretrained AI models (Buolamwini and Gebru, 2018; Nadeem et al, 2021), or examining how the data used in machine learning may lead to algorithmic biases against disabled people (Hutchinson et al, 2020). These inquiries have produced valuable insights about technical design of AI systems and methods to probe the types of bias exhibited by a system's parameters and output patterns. However, they often hold to a perspective that these biases are situated in the technical system and may thus be targeted with a technical "fix" via debiasing methods to produce supposedly unbiased AI (Bennett and Keyes, 2020). Our analysis illustrates how AI technologies cannot be divorced from the human contexts in which they are developed and used, and situates "AI bias" as a fundamentally sociotechnical issue emerging from situated practices that require sociotechnical action to mitigate.

The sociotechnical nature of algorithmic systems is the subject of a vibrant dialogue in the human-computer interaction (HCI) literature, and our discussion interacts with this dialogue in several key ways. As Whittaker et al (2019) and Alkhatib (2021) illustrate, algorithmic and AI systems are typically developed within existing (and inequitable) structures of power, and their design and operation reinforce these power structures and the injustices they already produce. Moreover, the ability to apply algorithmic systems at superhuman scale creates the potential to exacerbate these injustices, and the illusory presentation of algorithms as "objective" and therefore neutral creates a powerful narrative of impartiality that masks the subjective and highly contextualised *use* of the technologies (Whittaker et al, 2019; Alkhatib, 2021). For example, many mental health apps emphasise self-management approaches over contextual understanding in their design, deflecting attention away from structural injustices while presenting a veneer of technological empowerment (Weinberg, 2021).

These discussions have begun to interrogate the hidden narratives and ideologies of algorithmic and AI technologies. We deepen this line of inquiry by critically examining the interactions of disability with AI technology design, and using practical examples to investigate how disability ideology is digitally embodied in specific technical aspects of design and implementation. The extensive literature on AI and accessibility, which we highlight in more detail below, often brings a more technical emphasis to its analysis; our discussion dovetails with this literature by focusing on the use of AI in information and data processing, rather than in direct user interactions. We thus follow Noble (2018) and Eubanks (2018) in presenting a broader analysis of AI technologies that illustrates the interactions between social and organisational aspects and technical design, and situates the question of equity and AI as lying primarily within these interactions.

In terms of the disability literature, our article is aligned with a critical disability studies perspective (Garland-Thomson, 2013). We examine the AI development process from a crip technoscience perspective, which reflects "the co-production of science, technology, and political life" and regards disability as an important and desirable part of the world (Hamraie and Fritsch, 2019). Following the goals articulated by Meekosha and Shuttleworth (2009), we present a theoretical framework to interrogate often-unnoticed assumptions in the praxis of (disability-related) AI development. We illustrate the value of our framework by contrasting AI technology development under the medical model of disability with technologies developed under the social (Oliver, 1983) and relational (Kafer, 2013) models that have emerged from the last several decades of disability thought. Grounded in these ways of conceiving of disability, we show how divergent technological endpoints, combined with the lack of disabled people's input and perspectives in the AI development process, lead to both representational and allocational harms in social processes that use AI technologies.

To inform our analysis and reflect the deeply intersectional nature of disability, we draw on the robust methods that have been developed by critical digital race studies, queer studies, and Black technoscience scholars who challenge the interlocking networks of racism and other networks of oppression (see Noble, 2018; Benjamin, 2019; Hampton, 2021; McIlwain, 2019). For instance, Chun's (2011) articulation of race and/as technology interrogates the ways ICTs, such as AI, both work alongside racist ideologies to reinforce racism and can be used as a tool of racism themselves. Just as ableism is interconnected with racism, so ableism too is reinforced by technology and is used as a design tool (Williams et al, 2021; Spiel et al, 2020; Rauchberg, 2022). Birhane and Guest (2021) interrogate the ways in which the current computational science ecosystem creates many pressures, assumptions, and norms that oppress and exclude people of color and queer people. We argue that these observations and calls for decolonisation may easily be extended to include the ways in which disabled people are oppressed and excluded in computation (including AI). We do not see ableism as an ideology separate from racism, and the same critical tools that shed light on technological racism can inform our understanding and change of technological ableism. Moreover, just as racism, ableism, and homo/transphobia are inextricably entwined, so too must our critical tools enable an intersectional analysis. While we illustrate our framework using examples in disability, we hope that the questions we identify and methods we present can serve to interrogate many different kinds of oppression in AI and algorithmic technologies.

**The Need to Interrogate Ideology in AI Development**

Our focus in this article is not on how AI may support access needs or improve quality of life. Rather, we are concerned with the mechanisms through which ableist ideologies are often assumed and rarely interrogated in AI development, and the resulting ways in which AI technologies tend to engage in curative violence and contribute to the erasure of serve to erase disability as a cultural and political identity (Kim, 2017; Williams and Gilbert, 2019). AI and other information and communication technologies (ICTs) are typically designed in environments built on ableist structures and ideologies, which are sufficiently pervasive as to be normalised and left unquestioned. This can lead to AI technology development with goals that are well-intentioned on the surface, but reflect underlying ableism that actively harms disabled people. For example, AI technologies are often conceptualised as informing intervention-based cures that can help to "heal" or "erase" disability, e.g., by reducing the "deficit" between disabled people and the "normal" function which is perceived to be desired. This both stigmatises disability as "wrong" and in need of fixing and frames analysis in terms of impairments and pathologies rather than the disabled person's lived experience and opportunities to participate in different aspects of life and society (Stramondo, 2019; Williams and Gilbert, 2019; Kafer, 2013). Additionally, non-disabled creators and technologists bring top-down assumptions into the technology creation process, where disabled people are not agentic, but are rather treated as passive users testing out the ideas of designers who are almost always non-disabled (Gardner et al, 2021).

There are multiple interlocking mechanisms that support ableism in AI development. Disabled people are neither assumed to be nor hired to work as creators and designers of AI technologies, excluding them from having agency in the development and evaluation of AI technologies with direct impact on their lives. As Sloane et al (2020) show, efforts to improve different marginalised communities' "participation" in AI design often fail to make meaningful change in representation and input, as this participation is solicited in the form of external feedback rather than participatory design as a true collaboration from conceptualisation to implementation. This is intertwined with the curative goals that often drive conceptualisation and implementation of health-related AI technologies (i.e., that they will be used to help cure or fix the "harm" of disability), which are often based on normative assumptions of disability as deviation from a desired social norm. Moreover, disability is often treated as a monolith (e.g. seeking input from arbitrary "disabled" users, regardless of individual experience), which does not represent the multidimensionality of collective experience or practice. By drawing on the crip technoscience perspective articulated by Hamraie and Fritsch (2019), we reject the position that disabled people are only acceptable when they conform to a non-disabled world. Instead, we call for AI development processes that see and include disability—in all its multidimensional ways of being—as a desirable part of the world, and that recognise AI technology as inextricably political in materialising that view.

To develop a praxis for more just development of disability-related AI, it is important to first understand the patterns in how unjust AI harms disabled people. Assistive AI and other technologies, though motivated by principles of universal design, can nonetheless often place more onus on disabled people to generate accessibility, thus reinforcing ableism, assumed neurotypicality (i.e., the assumption that all brains should function in the same way), and audism (assumed superiority of hearing to Deafness) in technology creation. Ableism in AI development—which is often an unconscious process—is often rooted in assumptions of compulsory able-bodiedness and -mindness (McRuer, 2006; Price, 2011; Kafer, 2013), leading to these additional access-based burdens on disabled people both as technology users and as subjects of analysis. For example, sign language gloves and translation technologies tend to place the burden of access on Deaf people to conform to an ableist assumption of hearing as the norm, rather than shift and challenge communication norms to support multiple modalities (Erard, 2017; Bragg et al, 2019). Computer vision technologies for Blind people often undervalue the non-visual sense making skills of Blind people by relying on visual signal processing exclusively rather than multimodal integration with sound, etc., and can lead to further surveillance and privacy harms (Bennett & Keyes, 2019). Affective/emotional AI technologies for autistic people (particularly autistic children) assume normative expectations and views of emotion, and violently impose these norms on autistic people (Williams and Boyd, 2019). As Bennett and Keyes (2019) and Keyes (2020) discuss, the use of AI technologies to automate diagnosis of autism and other labels reinforce the legitimacy of existing power structures that remove the agency of the disabled people who are subjects of analysis. Such top-down and allistic, or non-autistic, approaches emphasise curative violence that erases autistic and intellectually/developmentally disabled (I/DD) users and creators (Williams et al, 2021; Rauchberg, 2022). These approaches, which fail to draw on extensive experience in engaging people with disabilities as partners in participatory design (Louw, 2017; Spencer González et al, 2020), wrongfully assume that I/DD people do not have agency in their insights on technology creation or user experiences, and reinforce the ableist idea that only non-disabled people are valued as technologists or users. Similarly, AI-powered hiring algorithms may fail to recognize disabled ways of living and working, furthering the exclusion of disabled people from the workforce (Kelly-Lyth, 2021; Tilmes, 2022; Equal Employment Opportunity Commission, 2022).

Many of the harms studied in prior work have occurred in situations where AI technologies are serving as user-oriented tools, with fewer studies investigating AI use in data analytics and decision-making. However, the patterns of exclusionary design and the use of technology to exert pressures to conform to societal norms, which underlie all of the examples cited here, are highly relevant to the use of AI for information processing in decision-making contexts, suggesting the need for further critical analysis in this arena. AI analytic technologies can contribute to medical harms (such as traumatic surgery or medication prescribed at the convenience of the medical system rather than based on a person's needs), social and financial harms (such as refusal of benefits or exclusion from the hiring process), and political harms (such as over-policing and disenfranchisement). We thus draw on both these examples and examples of access-related harms to raise the question: what does more just development of AI technologies look like? More specifically, how do we shift from a *disability* technoscience–grounded in assimilation, compulsory able-bodiedness, and designing for rather than design with–to a *crip and neuroqueer*[3] technoscience, rooted in multiplicity, collective access, and friction as technology?

Critical disability approaches to AI programming and research present new possibilities for agency for disabled researchers and users, in which their leadership and expertise are at the core of technological design. As defined by Williams et al (2021), crip human-computer interaction (HCI) "recognizes the researcher as situated, and thus articulated within, the sociotechnical meta-contexts of society, scholarship, research, design inquiry and practice" (p. 28) From the point of view of AI as a tool for information processing, we complement this study of interaction by drawing on Galloway's (2008) conception of computation as an "unworkable interface" whose very nature is its information loss: AI tools are inherently a palimpsest, erasing the multidimensionality of the individual person and replacing it with an approximation based on what is deemed "important". Thus, cripping AI development[4] for data analytics (as for user-oriented tools) requires both conceptual and practical change. We must create more space for disabled creators to participate and thrive in AI ecosystems, and to be equal partners and leaders in this development process. And we must decompose how AI information interfaces (in Galloway's sense) are designed, and map alternative paths to representing and encoding[5] the complex experiences and situated practices of disabled people in information and in information systems.

**Data analytics for decision-making: two example use cases**

Motivated by the need to better track the relationship between ideology in AI technology design and ableism in AI technology products, we structure our discussion around two example use cases to which we will apply our analytic framework. The umbrella of "AI technologies" includes both end user-facing technologies for accessibility and more "under the hood" uses of AI for information processing and data analytics, which often occur behind the scenes and may be invisible to end users. The role of AI technologies in accessibility is an active and growing area of study, particularly in the HCI literature. Current research foci in this area include developing and improving technologies for accessibility (Raja, 2016; Wu et al, 2020; Zhang et al, 2020), evaluating existing technologies through an accessibility lens (Kushalnagar et al, 2014; Gleason et al, 2019; Bennett et al, 2021), and critical analysis of the role of technologies in mediating accessibility (Ellcessor, 2016; Alper, 2017; Goggin, 2017; Miller, 2017; Shaheen and Lohnes Watulak, 2018). In this article, we complement these directions by focusing on "under-the-hood" use of AI for information processing, particularly AI technologies used to inform decision-making processes in healthcare, employment, and policy. Decision-making is a rapidly growing area of AI application and one with significant risks for inequitable development. Often, the people who are affected by AI-informed decision making have minimal participation in the development of the AI technologies that affect them. Moreover, the people affected by such AI technologies rarely have access to them or information about their design, and may even be unaware that AI technologies are being used in a way that affects them. Our work begins to lay a path towards disability-led design of AI technologies for decision-making settings, not only to improve the equitability of these technologies but also to imagine co-creating disability-led and anti-ableist AI technologies.

We structure our discussion of AI development for decision-making purposes using two example use cases set in two decision-making contexts that have historically pathologized and marginalised disabled people: healthcare and government benefits. We focus not only on the use of AI technologies to *make* (or recommend) decisions automatically, but also to *inform* decisions made by humans, often as one part of complex, pre-existing processes. Our example use cases are:

(1) In the healthcare setting, using AI technologies to analyse and combine information from disabled people and their healthcare providers about specific challenges and priorities for care. This may include a wide range of information, such as personal descriptions of long-term or daily life priorities, disabling situations or specific barriers, particular health conditions, availability of assistance and support, and more. Using AI-powered data analytics to bring all these disparate types of information together as part of the discourse of healthcare can inform people's experience of care, decisions about the course of care or specific treatments, and allocation of support resources such as assistive devices or home health aides.

(2) In the government benefits setting, using AI technologies to analyse applications and collected evidence to help assess eligibility for financial benefits under regulated programs. This may include identifying and analysing information about medical conditions, disabling situations and environmental barriers, relationships and support structures, and more, drawing on a range of personal, medical, and administrative data. While using AI technologies to make these high-impact eligibility decisions directly would be inappropriate (an issue we return to later), they could be of significant help in the process of reviewing and synthesising the available evidence to support a benefits application.

These use cases represent common scenarios encountered by disabled people, and key points at which the introduction of AI technologies for information processing and data analytics has the opportunity to both help and harm. On the one hand, AI technologies developed with a sensitivity to the situations, needs, and experiences of individual disabled people can help advance decision-making with that same sensitivity. On the other, top-down technologies that impose a particular worldview of disability can reinforce and worsen existing inequities. We examine how normative assumptions and a lack of disabled inclusion in the AI design process can lead to harmful technologies that perpetuate unjust power structures and erase the lived reality of disability in favour of a focus on "restoration" to non-disabled life—a form of what Kim (2017) describes as "curative violence." We lay out a basis for actively interrogating each aspect of information processing technology design to critically examine how AI technologies materialise and interact with disability. Our discussion thus serves as a first step towards illustrating what a disability-led and anti-ableist design process can look like for AI technologies in decision-making.

**Defining disability: An essential challenge**

To illustrate the impact of specific design decisions on an AI technology for data analytics, we draw on three conceptual models of disability that have played significant roles in policy, practice, and activism in North America and Western Europe in the twentieth century. We do not present these models and definitions to advocate for their use in AI development, nor to claim that any is sufficient on its own to represent disability experience. We also note that these are by no means the only models for conceptualising disability. Rather, these three models—the medical model, the social model, and the relational model—represent historical perspectives that continue to inform decision-making today, and we take them as examples and important reference points for examining how definitions of disability affect AI technologies in very practical ways. Each of these models presents a different perspective on what disability is, how it may be perceived or measured, and who or what is involved in producing disability identity. Our examination of them is thus intended as an illustration to facilitate the process of defining disability in a specific context for AI development. To accompany our discussion of the three selected models of disability, we provide example high-level descriptions in Table 1 of potential AI technologies that could be developed for our two use cases under each model, including descriptions of what information might be deemed relevant for analysis and how those AI technologies might be operationalised within their decision-making context.

**Table 1.** Example hypothetical purposes for developing AI technologies to support two use cases for decision-making related to disability under three different ways of defining, measuring, and operationalising disability. For each use case, we describe an example of how AI technologies might be used to process relevant information, and what the output of these technologies might be used to do within the decision-making process.

| Use Case | Model | What the technology might do | What it might be used for |
|---|---|---|---|
| Healthcare: collaborative priority-setting and decision-making | Medical model | Capture, monitor, and analyse information on body functions and structures to detect changes in health trajectory | Prioritise treatment targets and strategies based on patient's health priorities and provider goals |
| | Social model | Monitor and analyse physical environments to identify barriers to desired function and suggest appropriate assistive devices and/or accessibility changes | Prioritise therapeutic interventions based on person's priorities and barriers in functioning |
| | Relational model | Analyse information about access to health care, social support, and government resources to identify gaps in supporting the person's stated goals and priorities | Help to identify other disabled people with similar goals to connect and grow community and power |
| Government benefits: evidence review for eligibility assessment | Medical model | Analyse medical evidence to assess whether a person meets impairment-based eligibility criteria for benefits | Recommend additional medical screening to assess relevant criteria |
| | Social model | Analyse evidence of functional limitations and functional needs to assess whether benefits are needed/appropriate to cover the gap | Recommend additional self-reported data collection or professional assessment on functional status |
| | Relational model | Analyse evidence of needs for assistive resources and support structures to identify resource gaps that financial benefits can help allay | Recommend additional programs and services to connect the disabled person to regarding their non-financial needs |

The **medical model** frames disability as an attribute of a person, typically stemming from a particular health condition or injury. Medicalization of disability has facilitated significant and consistent harms to disabled people since the Industrial Revolution, from eugenics policies to medical errors and unnecessary interventions (Peña-Guzmán et al, 2019). Under a medicalized perspective, disability is defined in terms of meeting specific, primarily medical criteria—disability is perceived as something *diagnosable*, and is often equated with a medical diagnosis and thus directly representable in data. A person with muscular dystrophy or a lower extremity amputation is therefore seen as inherently *having a disability*. The medical model reduces disability to a specific "problem", which healthcare then aims to "fix." Social support programs operating on the medical model in government or civil society largely aim to provide financial or other resources to people based on perceived *capability* as a result of their disability, often regardless of differences in individual experience. Despite its harmful history, the medical model of disability remains the predominant operational definition in disability policy (Smith-Carrier et al, 2017), employment (Barnes and Mercer, 2005), and medical practice (Shakespeare et al, 2009) in much of the global North, and the medical model's interactions with the logics of colonialism have further informed much of how disability is conceptualised within power structures in the global South (Grech, 2015). The medical model therefore serves for many people around the world as the default understanding of how "disability" is defined in daily life. Under the medical model, AI technologies are thus primarily developed to analyse information related to diagnosis and pathology, ignoring the personal experiences and dynamic contexts of disabled life and representing disability as a collection of biological or biomechanical ills to be "cured". The use of these technologies in turn emphasises curative treatments and goals rather than person-centred interventions that are sensitive to an individual's lived experience and needs.

The **social model of disability**, by contrast, conceptualises disability as a phenomenon emerging from the interaction between a person and their environment. On this view, disability is neither static nor internal as in the medical model: rather it is dynamic and external, emerging from situations and environments that contribute to a process of enablement or disablement (Verbrugge and Jette, 1994; Shakespeare, 2006*)*. The social model thus in principle requires measuring and understanding both a person's capacities and needs (physical, cognitive, and otherwise) with respect to functioning in different activities and social roles, and the facilitators and barriers to that functioning presented by a given environment.

The social model has its roots in sociological research in the mid-20[th] century (Nagi, 1965) and since its formal articulation in the 1980s (Oliver, 1983) it has become one of the dominant academic perspectives across disability studies, social policy research, and rehabilitation science. The World Health Organization's International Classification of Functioning, Disability and Health (ICF; World Health Organization, 2001), an internationally-accepted standard for defining and describing disability, draws primarily on the social perspective in its biopsychosocial model of human function, and disability policies in the United Kingdom and in Scandinavian nations draw on the social model's focus on contextual factors (including physical, social, and cultural environments) (Mercer and Barnes, 2004; Christensen et al, 2008; Lindqvist and Lamichhane, 2019). However, the multidimensional nature of the social model has proven difficult to operationalise in practice, and measurement and decision-making in healthcare, employment policy, and other areas often still rely primarily on medical definitions, in part due to the ways these definitions have been built directly into bureaucratic structures and processes (Roulstone, 2004; Bingham et al, 2013). The social model has also been criticised by disability studies scholars and others for overreliance on environmental factors, leading to de-emphasis and de-valuing of biological and medical factors that affect people's health and personal experience (Oliver, 2013; Owens, 2015; Shakespeare, 2017). The social model has been used to guide only limited development of AI technologies, but these have focused primarily on information about experienced limitations in function and barriers to function from a whole-person perspective, as opposed to focusing on specific body structures or body functions (Agarronik et al, 2020; Newman-Griffis et al, 2021).

The third model we draw on is the **political/relational** model of disability (cf. Kafer, 2013). This model (which we refer to as the *relational model* for the remainder of the article) bridges some aspects of both the social and medical models, but in contrast to their definition of disability in terms of individual experience, the relational model frames disability as "experienced in and through relationships; it does not occur in isolation" (ibid. p8). The relational model emphasises the role of disability as a political identity and therefore a site of collective action, and as a dynamic category emerging from and affecting social relations. The relational model has emerged in part together with the growth of critical disability studies as a field over the last two decades, but it is strongly rooted in the history of disability justice and crip activism (primarily in North America) in the twentieth and twenty-first centuries that has helped build an academic and policy understanding of the intersectionality of disability with other marginalised identities and experiences (for an overview of the history and intersectionality of the disability justice movement, see Project LETS, 2021). For instance, Sins Invalid's (2019) disability justice primer articulates how information and communication technologies, such as the Internet, can be hubs for disability justice organising and activist practice (ibid. p. 25). Coalitional groups like #PowerToLive and #NoBodyIsDisposable use social media technologies to advocate for offline social change around racial, health, disability, and environmental justice (#PowerToLive, n.d.; #NoBodyIsDisposable, 2020). Moreover, Jackson et al.'s (2022) conceptualization of disability dongles takes an activist approach to challenging disabled peoples' lacking agency in the creation, design, and use of assistive technologies that do not support a user's access needs, but instead create AI with an aim to "cure" disability.

Though we do not fully take a disability justice approach in this paper, disability justice practice shapes our understandings about what a political/relational model brings toward investigating the relationship between disability, political identity (e.g., race, gender, sexuality, class), and technology. On this view, disability is conceived of as a shared political experience and becomes a way of identifying with and relating to others for community-building, support, and inclusion;

as well as a way of categorising others for marginalisation, and exclusion. The relational framework also emphasises interdependence in thinking about disability, reflecting the ways in which access in all senses is affected by community support structures. A relational perspective on disability thus critically extends the social model to account for collective factors of access, support structures, governance, and political power. The relational model facilitates analysis of the interaction between individual and collective needs, and the ways in which political power to address these needs is deeply rooted in community identity and structures. Analysing technology from a relational perspective therefore centres the political aspects of technology development, in terms of the exchange of ideas, priorities, and decisions between different parties involved in the development process—a group which may or may not directly include disabled people. A relational analysis also resists the depoliticization of disability through a narrative of technological neutrality or impartiality by showing that technologies are produced by people, and algorithms necessarily advance the ways of perceiving the world held by their designers. The relational understanding—that because disability is inherently political, so too must be anything that materialises disability, including information technology—is central to our critique of AI development practices. While no AI technologies we are aware of have been developed with an explicit alignment to the relational model, Table 1 imagines possible goals for AI technologies focusing on information about power and support structures, communal resources, and building interdependent connections.

*Interactions between these definitions*

These three models, and other definitions of disability not described here, are neither mutually exclusive nor strictly complementary. Rather, each highlights different aspects of individual experience and health as more or less salient, and prioritises different factors in defining what is and is not disability or being a disabled person. It is in the operationalization of these definitions that they may come into conflict. For example, a person with a medical diagnosis of chronic pain may meet medical criteria of disability, but with access to appropriate medication and support structures may experience minimal disablement from a social model perspective. Similarly, a person with severe arthritis may experience considerable disablement in environments that require high mobility, but may not be diagnosed with a medical condition that meets the political/formal definitions of disability and will thus lack access to helpful state-provided financial, social, or medical resources. Disability is multidimensional, and defining what disability might mean and how it might be materialised in information technologies is not a singular process.

When new technologies are introduced into processes that involve making decisions about disability—whether or not those technologies are explicitly disability-related—the presence and use of the technology affects both the behaviour of decision-makers and the ways in which their decisions affect disability identity. These technologies thus become both part of the environmental context of the social model (by affecting the sociopolitical processes disabled people engage with) and the personal and political context of the relational model (by affecting the production of disability identity and who is involved in that process) (Forlano, 2017). Even if information technologies are initially conceptualised under one specific model at the design stage, these technologies must therefore be critiqued through multiple perspectives to understand the impact of their stated goals and the effects of their use.

**Framework for critical examination of AI analytics and disability**

Bias is neither inherent nor ineffable in AI technologies: it emerges as a result of specific elements and decisions in the process of designing AI systems. With the components of our analysis in place, we have what we need to discuss key elements and decisions in AI problem formulation, and use them to understand how drawing on different models of disability in our example use cases can produce AI technologies with a range of different biases. The definition of disability used by the designers of an AI technology, or the omission of any explicit consideration of disability, will affect the role that technology is intended to play in decision making, the data it will analyse, its algorithms, and its output. Different definitions of disability will change not only the operation of an AI technology, but the worldview of disability that it serves to reinforce in the broader context it is used in. Critical tools to identify and define the specific ways in which disability is conceived, measured, and implemented within an AI technology can therefore inform more responsible technology design, evaluation, and management.

AI technologies for data analysis are broadly intended to take some information on a person or situation in the world and distil it into some insight, pattern, or subset of information that supports making a decision about that person or situation. To design an AI technology to address a specific information problem, the first step is defining a situated *problem formulation* (a terminology we adopt from Obermeyer et al. (2019)). A problem formulation has several different components, each of which require making distinct design decisions in practice. In our analysis, we examine four fundamental elements of an AI problem formulation:

1) The overall *scope* of the technology within the specific situation(s) it is designed to address;

2) The *data* that the technology will be used to analyse, i.e. the information available about the person or situation;

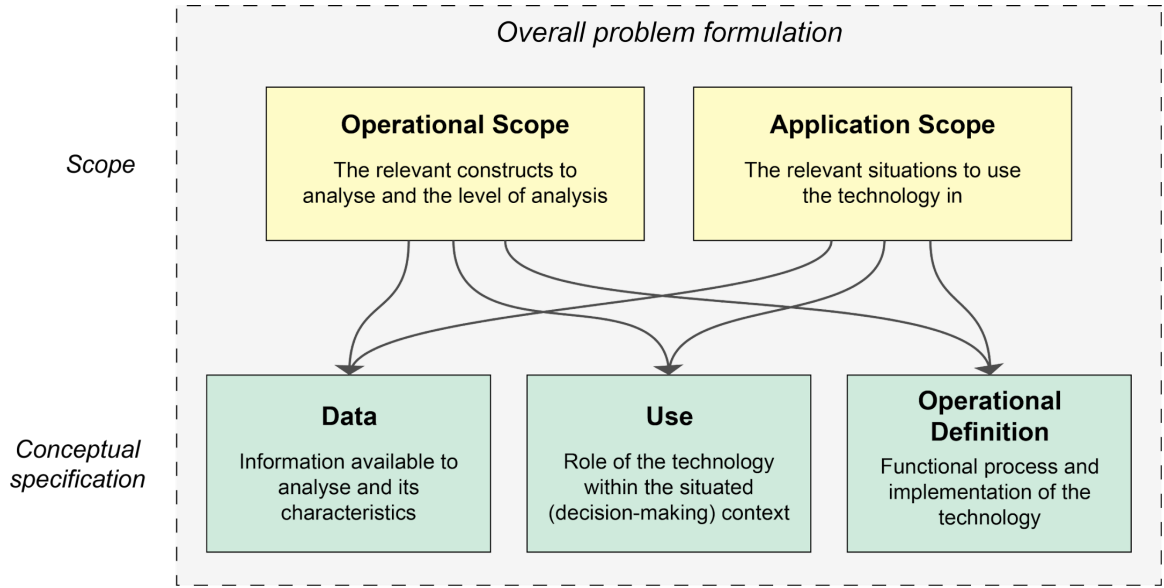

**Figure 1.** Conceptual illustration of the elements of AI problem formulation for data analytics. We separate the components of operational and application scope, which form the overall scope of the technology's use, and the components of data, use, and operational definition, which form the conceptual specification the technology is meant to implement. Image Description: A diagram of overall problem formulation, broken up into two levels of boxes for the different components of problem formulation. The first level, Scope, includes boxes for Operational Scope and Application Scope. The second level, Conceptual Specification, includes boxes for Data, Use, and Operational Definition. Arrows connect the Scope boxes to the Conceptual Specification boxes to show directions of impact.

3) The *use* the technology will be put to, i.e. its role in the broader decision-making process;

4) The *operational definition* of the technology, i.e. the algorithmic and data operations by which it uses the data to address the problem.

As illustrated in Figure 1, the data, use, and operational definition for a technology form an informal conceptual specification, which interacts with the technology's scope to define a situated problem formulation.

To make our discussion more concrete in illustrating these elements of AI development and their interactions, we will consider one type of scenario in which disabled people might come into contact with AI technologies that have been developed for our two use cases in healthcare and government benefits. We use as our illustrative example imagined experiences of people with spinal cord injuries who use mobility aids (e.g. wheelchairs). To explore a variety of data types and sources, we assume that our population has access to a wide range of healthcare and social services, that many of them use self-reporting instruments and tools to provide details on their own priorities and experiences to other decision makers, and that some (though not all) are using environmental sensors and/or video recording devices to capture information about their physical and social environments. We do not consider any of these data sources to be more or less valid than any other, nor do we expect any particular type of data to be recorded or available to inform AI development. Rather, we have chosen these data sources to illustrate the range of data that could be used to inform AI systems and discuss how different kinds of information interact with different definitions of disability. Similarly, our focus on mobility-related disability is chosen simply as a common example within which to analyse AI development. Investigating how the parameters and prioritisation of AI design change to reflect the diverse dimensions of disability is an important area of further research to build on our initial framework.

As the scope governs the specific decisions made about data, use, and operational definition in practice, we first describe the components of setting a technology's scope and then examine the elements of data, use, and operational definition for two example technology scopes within our illustrative scenario.

*Initial problem formulation: Setting the scope for the technology*

The first component of design to analyse is the problem definition an AI technology is designed to address. While this may seem straightforward, the process to translate a broad goal for technology development into an actionable problem formulation requires identifying and describing the particular context in which a technology will be used, and making specific decisions about what information is relevant to analyse and how. Each of these steps then becomes a jumping-off point for different conceptualisations of disability to lead in distinct directions of technology formulation.

The first step in the process is to define the scope of the analytic technology, along two dimensions: the operational scope and the application scope. The operational scope is, in effect, how specific or generalisable the technology is designed to be. Considering the healthcare use case in our scenario, an AI tool to identify key concepts and evidence in an individual note from a single encounter with health or social services is of narrower scope and more generalisable (i.e. can be used in a wider variety of applications) than a tool designed to synthesise longitudinal information from regular physical therapy encounters into a personal timeline. The application scope, then, describes the range of situations in which the specific technology being developed is intended to be used. For example, the AI tool to identify key concepts and evidence in a single note (which has a narrow operational scope) may be intended for use only by physical therapists (a narrow application scope) or may be intended for use in analysing the records from all healthcare and social services encounters a person has (a broad application scope). It is important to note that the scope in which a technology is used *after it is developed* may change over time— for example, a tool for use by physical therapists may be adopted by social workers. This type of scope creep or shift in purpose is a key challenge for future work on extending our framework beyond the design stage into the management of AI technologies over time.

The operational scope and application scope interact with definitions of disability primarily around determining what is *relevant*. The operational scope of an analytic technology, as we have defined it, involves identifying the relevant *constructs* the technology will be designed to analyse. Taking our example of identifying key concepts and evidence in an individual document (narrow operational scope in our healthcare use case), the "key" concepts to target might include medical findings and treatments under the medical model, environmental factors and functional limitations under the social model, or people and services under the relational model. Setting an operational scope requires asking, in effect, what counts as information of interest based on the purpose we (as designers) aim to address with the technology and how we conceive of disability. The application scope, similarly, involves identifying the relevant *settings* in which the technology will be used for analysis. Again considering our narrow scope example of analysing only one type of documents, we may target primary care records under the medical model, occupational therapy records under the social model, or social work records under the relational model.

A design team's definition of disability alone does not directly determine the operational or application scope of an AI technology, and the examples given here are not comprehensive in terms of the variety of decisions that might be made. Our concern is to illustrate the role of disability definition as a key factor influencing the problem formulation step: it is one of the primary factors in translating a purely technical idea (single pieces of evidence, multiple records, etc) into a situated description of the purpose of the technology (single medical findings vs environmental factors; multiple healthcare records vs cross-cutting personal, social, and medical information).

Once operational and application scope are identified, determining the conceptual specification of how they are implemented in practice requires understanding the three interconnected elements of data, use, and operational definition. Each of these interacts with definitions of disability in different ways to produce very different technologies for any particular broad goal and target context. For concreteness in discussing each of these aspects of design, we will further specify our example use cases with the following operational and application scope (leaving the specific definition of "relevant" open to examine the effects of using different models of disability):

*Healthcare use case:* Identifying relevant outcomes/status and the evidence that supports them across the complete collection of data available for a person (including personal reports, sensor data, and health and social services records). The outcomes will be presented to our example disabled population and healthcare providers during consultations, to inform priority-setting and decision making about next actions.

*Government benefits use case:* Identifying relevant pieces of evidence across the complete collection of data available for a person, in terms of assessing eligibility for financial benefits. The pieces of evidence will be presented to benefits assessors with links back to the supporting data, to inform the eligibility assessment process.

Table 2 illustrates different possible ways of selecting data, the use of the technology within its context, and an operational definition for each of these examples.

*Data: What is available to analyse*

The first component of specifying a problem formulation within a given scope is identifying the data that are collected within the application scope, and therefore are available for use in data analytics. Beyond just *what* is collected–as well as what is *not* collected but would be informative to know–understanding *why* and *how* information is collected is also fundamental to both responsible and effective problem formulation. There is extensive literature on assessing different methods of measurement and data collection; we briefly highlight as a starting point three initial questions to guide informed review of the data available for analysis.

<u>For whom is the data collected?</u> In our example scenario focusing on users of mobility aids, if information on curb cuts and stair-only access in one's home neighbourhood is only collected for people who are receiving government disability benefits, then this information is of limited value for our healthcare and government benefits use cases. In the healthcare use case, it would only represent a subset of the population and so should not be relied on as a population-level indicator; in the government benefits use case, this information would not even be available for initial determination of benefits and is therefore inappropriate to use in designing data analytics for initial determination (cf. Eubanks, 2018 for a discussion of similar issues in technological systems in welfare contexts).

<u>Who records the data, and what perspective do they bring?</u> If information on the personal priorities for daily living of our example population is only recorded by healthcare providers asking specific questions, then this information may not be representative of what they would say for themselves in self-reported data. Similarly, data collectors have their own positionality separate from that of the technology used to analyse their data: data from medical encounters will generally tend towards a medical view of disability, while self-reported data will often focus more on relational aspects and personal experience.

**Table 2.** Examples of how the same goal and scope for a data analytic technology can yield very different detailed problem formulations for analytics when filtered through different definitions of disability. We examine the elements of conceptual specification within an overall problem formulation, in terms of (1) the data available for analysis; (2) the operational definition of the analytic technology; and (3) the role it is intended to play in decision-making. We show illustrative descriptions of each of these elements for two use cases: collaborative priority-setting and decision-making in healthcare, and evidence review for eligibility assessment in government benefits programs.

| Technology scope | Disability definition | Aspects of problem formulation | | |
| --- | --- | --- | --- | --- |
| | | *Data* | *Operational definition* | *Use in decision-making* |
| Healthcare: Identifying relevant outcomes/status and the evidence that supports them across data types, to inform priority-setting and decision making about next actions. | Medical model | Medical records, in-home sensors, wearable devices, self-reports of physiological or cognitive conditions | Analyse patterns in numeric and spatiotemporal data to identify persistent/severe health conditions and impairments. Analyse text reports for mentions of such conditions and impairments. | Connect subjective reports to objective evidence of impairment to support treatment and rehabilitation decisions |
| | Social model | Self-reports of experienced limitations and barriers, environmental sensors, images and video, infrastructure and health and safety reports, medical records, in-home sensors, wearable devices | Analyse reports for descriptions of environmental barriers and facilitators and functional outcomes. Analyse environmental data and medical evidence for specific characteristics of these barriers/facilitators. | Connect personal reports to details of specific environmental barriers and facilitators to inform use of assistive devices, pursuit of environmental access improvements |
| | Relational model | Health services records, social services records, evidence from advocacy organisations, self-reports of available resources and support systems | Analyse service usage records for patterns of access and exclusion. Analyse written reports for descriptions of unaddressed needs and enabling/disabling systems. | Connect person's needs to appropriate services and support systems to address gaps in needed resources |
| Government benefits: Identifying relevant pieces of evidence across the | Medical model | Medical records, application forms, commissioned expert assessments | Analyse collected reports to identify descriptions of medical impairments. | Match medical evidence to regulatory criteria for disability benefits |

| Technology scope | Disability definition | Aspects of problem formulation | | |
|---|---|---|---|---|
| | | *Data* | *Operational definition* | *Use in decision-making* |
| complete collection of data available for a person, in terms of assessing eligibility for financial benefits. | Social model | Professional assessments of functional activity and participation, in-home and environmental sensors, medical records, self-reports of limitations and barriers | Analyse text reports and environmental data to identify environmental barriers to work-related/daily life functioning | Link personal descriptions, professional reports, and objective evidence of functional barriers to needs for functioning in society to inform need for additional benefits |
| | Political/ relational model | Health services records, social services records, social work assessments, self-reports of available resources and support systems | Analyse service accessibility and usage and descriptions of available resources to identify resource needs for work-related/daily life functioning | Connect evidence of available resources to resource needs to assess need for additional support through financial benefits |

<u>What are the data proxies for?</u> Related to the perspective of data collectors, many data variables are collected through processes that themselves encode implicit definitions of what is relevant. As an example, physical therapy records for our example population may include categorisation of each person into groups based on a clinician's judgement of how intensive an intervention the person would benefit from. This reflects a medically-oriented decision-making process; an item on a self-reported questionnaire about experiencing difficulty doing activities around the home reflects an assessment under the social model. These variables, which are often used as targets for machine learning, must therefore be assessed for the conceptualisation of disability they may implicitly encode.

The data that are available for analysis are characteristics of the application scope, and are not directly affected by the definition of disability a technology is developed under. The selection of which data will be used by that technology, however, depends heavily on how disability—and its measurement—is defined by the technology designers. An AI tool developed to align with the medical model will focus on medical data; a tool developed to align with the relational model might use some of the same data sources, but will likely focus on very different kinds of information to be gleaned from them (as illustrated in Table 2).

*Use: What role the technology will play*

The second component of specifying the problem formulation is defining how the technology will be used in the decision-making process. This encompasses a wide range of questions, such as: Who is using the technology to gain insight from information, and who is using that insight to guide decision-making? (These are not necessarily the same person, for example a data analyst may use AI technologies to produce reports that inform the decision made by a benefits program adjudicator.) What other sources of information outside the scope for data analytics (e.g., policy considerations, quotas, limited available resources for allocation, etc.), affect that decision? When (and for whom) is the technology used, and what process, technological, or other controls are in place to govern the use of the technology?

These are complex questions whose answers often change and develop over time, and a complete accounting for them at any single point in time is rarely realistic. Nonetheless, a clear understanding not only of what a data analytic technology is *meant to do*, but also of how it *will (or could) be used* is a critical component of both design and evaluation. For example, technology designers may choose to focus on higher coverage of relevant information or higher confidence that what is produced is relevant—a common and frequently unavoidable tradeoff in AI development between precision (the proportion of what the AI system produced that it should have produced) and recall (the proportion of what the AI system should have produced that it in fact did). In our example scenario, if the technology is to be used to support human review of benefits claims, higher coverage of relevant information (e.g. about specific assistive devices, aspects of the built environment) may be preferable at the expense of erroneously identifying some information as relevant, knowing that it is easier for a human reviewer to filter out irrelevant results than to identify missing ones (Newman-Griffis and Fosler-Lussier, 2019). If the technology were to be used for automated cohort identification in population health research, where missing some relevant population members may be preferable to including people outside the target population, technology designers may instead focus on precision to minimise erroneous inclusion. In all cases, the designer's operating definition of disability affects the types of target contexts a technology may be appropriate to and what role it can reasonably fulfil within those contexts.

*Operational definition: How the technology will work*

The third component, and the focus of most design effort in current AI practice, is the operational definition of an analytic technology. The operational definition specifies how a technology will accomplish the use it is designed for: i.e. given the data that are available for analysis, how are those data to be processed to inform the role the technology is meant to play? Answering this question involves taking the constructs of interest (i.e. the operational scope) and the relevant data available and defining processes to identify/analyse those particular constructs within those particular data. There are three primary aspects to this definition: identifying what each part of the relevant data says about the target constructs, defining what the technology is intended to output from these data (in other words, what the technology is meant to tell the user about the relevant constructs), and defining the algorithmic process to get from input data to output.

These are very concrete questions, and the effects on them of the overarching model of disability used in the design process are largely mediated by the earlier steps of operational scope, data assessment, and defining the use of the technology. Nonetheless, it is at the operational level that the chosen definition of disability is most directly embodied in data analysis. Taking our healthcare use case as examined in Table 2, we see that under the medical model the data deemed relevant may include text reports, numeric measurements (e.g. from medical assessment), and spatiotemporal data. Since the constructs of interest are primarily medical conditions and impairments, the operational definition of the technology is to analyse the numeric measurements and spatiotemporal data for known patterns that correlate with specific medical conditions and impairments, and to analyse the text reports for mentions of these conditions and impairments. Under the relational model in this example, the constructs of interest are primarily access to services and other structural resources (including community resources): the operational definition of the technology is therefore to analyse service usage data to see what services a person has access to or is excluded from, and analyse textual reports they provided to extract information about their community support and other support structures.

The actual algorithms used to perform this operational definition, as well as specific technical decisions about model architectures and machine learning strategies, provide further points of departure for how a theoretical conceptualisation of disability may be embodied into an implemented technology. While an investigation of these more technical aspects of design is outside the scope of this article, using our framework to understand the context in which those implementation decisions are made can contribute significantly to making and evaluating those decisions from an informed sociotechnical starting point.

*How this framework can be used in practice*

Our critical framework is intended to serve as a starting place and a guide for discussion when formulating a new application of AI technologies, or when assessing one that has already been developed. Our framework by no means addresses all aspects of design, nor is it intended to constrain the discussion around how an AI technology should work and for whom. Rather, our analysis of the fundamental elements of AI problem formulation is meant to elicit unspoken assumptions and perspectives held by AI designers, and to provide specific points to facilitate reflection and discussion for drawing on multiple perspectives. We envision our framework being used by AI development teams (including disabled developers) as a structured tool to establish a shared understanding of the purpose of a disability-related AI technology and how its design will reflect that purpose. We also envision its use by critical and ethical scholars to deconstruct the elements of how an AI technology operates, and to identify specific decisions that contribute to a system's behaviour (desirable or undesirable). More broadly, our framework is intended to provide structure and specificity to the often nebulous process of deciding what an AI technology is supposed to do, and to provide grounds for disputing, debating, and coming to consensus around each of those decisions.

**Towards a Design Praxis for Disability-Related AI Analytics**

Every component of formulating data analytics technologies for disability is affected by the lens or lenses through which disability is framed. Our critical framework provides initial questions and considerations for examining how AI analytics technologies are designed, and a structure to guide efforts to design new technologies with a critical understanding of how they materialise and relate to disability.

Three themes will be key in building on this framework to develop a more robust design praxis for disability-related AI. The first is to dig deeper into the analytic design process, and to connect specific technical decisions back out to their conceptualisation and impact under different models of disability. Implementation details such as preprocessing input data to prepare it for analysis, algorithm and model architecture choices in machine learning, and data structures and representation of outputs serve as further ways in which ideologies of disability may be materialised through the details of technical design, and these too must be subject to critical examination. Extending our framework for these inquiries will help articulate the connections between critical analysis and technical implementation, and can serve as the basis for developing reporting standards and assessment criteria for design decisions in new technologies. Technical extensions of our framework will further help provide the scaffolding for developing a vision for multidimensional AI analytics for disability that draws on multiple ways of conceiving of, defining, and operationalising disability.

The second theme is one of breadth: improved identification of the sociotechnical aspects of the context surrounding technology use that affect the efficacy and power dynamics of AI technologies for disability, and whether they are appropriate to develop in the first place. For example, if the data available for analysis in a particular healthcare context do not include information about occupational health, personal or social environments, or community structures, AI technologies cannot realistically be developed under the social or relational models without first changing the processes by which data are collected. Entwined with the questions of appropriate data is the issue of appropriate purpose for a technology, and what the risks might be of using it or not using it. This was well-illustrated by the use of algorithmic approaches to automatically expand the list of people recommended to shelter at home in the United Kingdom due to COVID-19 risk: automated identification of vulnerable members of the population offered improved safety with respect to health, but also increased risk of physical and mental harm from isolation and restrictions in opportunities to participate in the workplace (Patel, 2021). There are higher level questions of fit that must be addressed as part of investigating appropriateness, in terms of who is driving the development process and why: for example, if the development of an analytic technology for a government benefits use case is being funded by an agency that subscribes to the medical model, they may be unwilling to accept a technology designed to refute harmful normative assumptions about disabled people. The transparency with which AI technologies will be implemented and used must also be examined: in many cases, people whose lives are affected by AI systems may not be made aware that those systems have been adopted, and may be excluded from information or agency on how they might be impacted by their use. These are particular risks for disabled people, who are often already excluded from participation in or information about many decision-making processes where AI technologies are being adopted.

The final theme is one of outward change of both power and practice in AI development. In addition to a better understanding of the representational and allocational harms that result from under-informed design of AI analytics in the disability context, the fundamental issue must be explored that *not all information about disability is equal*. Information from disabled people, who can speak most directly and accurately to their own lived experience, is often dispreferred or entirely ignored in favour of information from more privileged sources such as healthcare professionals or government staff, who may have little understanding of—or actively pathologize—disabled experience. In investigating how these issues interact with the AI technologies that are used to analyse collected information, it will be particularly useful to draw on the concepts of testimonial and hermeneutical injustice described by Fricker (2007), which reflect on the ethical harms produced when people with marginalised experiences are both devalued and underrepresented in design processes. Such an understanding can further help in assessing and improving access for disabled people to meaningfully contribute to the analytic design and evaluation process.

As a stepping stone to begin shifting power dynamics, we call on future work to critically include (and compensate) disabled people in the development of AI technologies. While structural changes are needed to achieve anti-ableism and end disability-based oppression in sociotechnical systems, inclusion and reframing technology development in terms of actively interrogating conceptualisations of disability may serve as a first step to imagining accessible and just futures where disabled peoples' expertise and agency is protected and amplified. As part of this interrogation of disability, it is essential as well to build on a framework of collective access (Hamraie and Fritsch, 2019) to recognise the multidimensionality of disability, and that what works for one group or one disabled person is not universal.

It is important to note that people with disabilities are *already* engaged in the (often unpaid) process of building and designing AI systems (Bigham and Carrington, 2018). For example, Blind people often undertake the unpaid labour of repairing screen reader errors through bug reports; Arab VoiceOver users curate datasets to correct misrecognition of Arabic words (Alharbi et al, 2022). However, the contributions of disabled people are largely dismissed from professional spaces (Bennett & Rosner, 2019; Ymous et al., 2020; Jackson, Haagaard, & Williams, 2022). Currently, AI systems are built by mainly non-disabled researchers who largely hold medical views of disability and imagine assistive AI as a tool for cure or rehabilitation (Williams et al., 2021; Ymous et al., 2020). Moreover, mainstream co-design practices involving non-disabled research teams often treat disabled people as passive "testers," reproducing ableism's structural violence (Rauchberg, 2022).

Our paper illustrates the importance of a disability-led approach, rather than one in which disability is incidental to design, by drawing on three key historical models of disability (medical, social, and relational) to illustrate how different ways of defining disability result in highly distinct AI technologies. While some models may lead to the reproduction of technoableism (Shew, 2020), others can provide hope and opportunity towards a future where ableism is eradicated and disability is celebrated. In addition to identifying appropriate models, we urge AI and data practitioners to critically co-create systems *with* (rather than for) people with disabilities. Genuine collaborations amplify the agency, leadership, and expertise of disabled creators and users, instead of pacifying representations that do not mitigate (techno)ableism's structural harms (cf. Hamraie & Fritsch, 2019). In doing so, underlying harms and assumptions in AI and data technologies can be made apparent and repaired. In HCI, accessibility scholars have begun centering the perspectives of people with disabilities from early stages of emerging AI technologies, surfacing unique considerations that are misaligned with popular AI research (Alharbi et al., 2022; Brewer and Kameswaran, 2018; Theodorou et al., 2021). While these approaches may not solve structural issues that mediate and amplify AI harms on their own, they act as an invaluable first step towards rectifying power imbalances and an important example to draw on in further work.

**Conclusion**

The proliferation of artificial intelligence (AI) technologies as behind the scenes tools to support decision-making processes presents significant risks of harm for disabled people. The unspoken assumptions and unquestioned preconceptions that inform AI technology development can serve as mechanisms of bias, building the base problem formulation that guides a technology on reductive and harmful conceptualisations of disability. As we have shown, even when developing AI technologies to address the same overall goal, different definitions of disability can yield highly distinct analytic technologies that reflect contrasting, frequently incompatible decisions in the information to analyse, what analytic process to use, and what the end product of analysis will be. Here we have presented an initial framework to support critical examination of specific design elements in the formulation of AI technologies for data analytics, as a tool to examine the definitions of disability used in their design and the resulting impacts on the technology. We drew on three important historical models of disability that form common foundations for policy, practice, and personal experience today—the medical, social, and relational models—and two use cases in healthcare and government benefits to illustrate how different ways of conceiving of disability can yield technologies that contrast and conflict with one another, creating distinct risks for harm.

Critical examination of disability-related AI technology development is not only crucial in guiding the initial process of developing new analytic technologies, but also in capturing and assessing the range of decisions that affected an existing technology and imagining alternative designs. This article provides both a starting point for that examination and a roadmap to strengthen and expand the critical tools available for understanding the relationship between AI technologies and disability.

**Notes**

[1] We use the identity-focused phrase "disabled people" and the person-centred "people with disabilities" interchangeably throughout this article, to indicate the nature of disability as both a category socially assigned to people and a personal identity expressed by people. For a deeper discussion of person-first and identity-first language, see Dunn and Andrews (2015).

[2] We use "ableism" as a blanket term to refer to both discrimination *against* people perceived or identified as having a disability (referred to in some contexts as "disablism" (Miller et al, 2004; Gappmayer, 2021)) and discrimination *in favour* of people perceived of identified as non-disabled.

[3] We follow Nick Walker's (2022) working definition of neuroqueer, which originated as '... any individual whose identity, selfhood, gender performance, and/or neurocognitive style have in some way been shaped by their engagement in practices of neuroqueering, regardless of what gender, sexual orientation, or style of neurocognitive functioning they may have been born with.' We also align our analysis with the 8th tenet of Walker's definition: 'Working to transform social and cultural environments in order to create spaces and communities – and ultimately a society – in which [neuroqueer]... practices [are] permitted, accepted, supported, and encouraged'".

[4] We here follow the field of crip theory (cf. McRuer, 2006) in using "crip" as a verb in addition to an adjective. "Crip AI" may be thought of as AI that sees disability as a desirable part of the world; "to crip AI" is then making this perspective shift happen within the AI research, development, and application communities. For an introduction to crip AI design beyond the decision-making setting, see Hickman (2021).

[5] We use "representing and encoding" here in a primarily technical sense: i.e. data and information as a (necessarily limited) way of conceptualising and measuring individual experience. There are, however, important questions to consider regarding broader senses of "representation" in re technology and AI development. Confronting assumptions that disabled people are not already present in technology development; creating space and safety for disability in the AI workplace and research community; curating disability-led spaces within AI technology and community; all of these are essential challenges to changing the broader landscape of AI and disability. While these issues are outside the scope of this particular discussion, we highlight them as key directions for building on the ideas outlined here.

**Acknowledgments**

This research was supported in part by the National Library of Medicine of the U.S. National Institutes of Health under award number T15LM007059.

**About the Authors**

Denis Newman-Griffis (they/them) is a Lecturer in Data Science at the University of Sheffield, Information School. Their current research interests focus on translational design and evaluation of natural language processing and artificial intelligence technologies for health and disability, with special attention to ethics and equity. Part of this work was completed while they were a National Library of Medicine Postdoctoral Fellow at the University of Pittsburgh.

Jess Rauchberg is a doctoral candidate in the Department of Communication Studies and Media Arts and at McMaster University, where she specializes in cultural studies of human-computer interaction, disability studies, philosophy of communication, and new media research.

Rahaf Alharbi is a Ph.D. student in the School of Information at the University of Michigan. Her research sits at the intersection of disability studies, accessibility and human-computer interaction (HCI). Working towards an accessible and just future, she explores ways to design and refuse emerging technologies with disabled people.

Louise Hickman is a Research Associate at the Minderoo Centre for Technology and Democracy at the University of Cambridge. She is an activist and scholar of communication, and uses ethnographic, archival, and theoretical approaches to consider how access is produced for disabled people.

Harry Hochheiser is an Associate Professor of Biomedical Informatics in the University of Pittsburgh School of Medicine. His research interests include human-computer interaction, information visualization, bioinformatics, universal usability, security, privacy, and public policy implications of computing systems.

*Definition drives design: Disability models and mechanisms of bias in AI technologies*

*Definition drives design: Disability models and mechanisms of bias in AI technologies*11. R. N. Brewer and V. Kameswaran, 2018. "Understanding the Power of Control in Autonomous Vehicles for People with Vision Impairment." *Proceedings of the 20th International ACM SIGACCESS Conference on Computers and Accessibility*. New York: ACM Press, pp. 185–197. doi: https://doi.org/10.1145/3234695.3236347

12. J. Buolamwini, and T. Gebru, 2018. "Gender Shades: Intersectional Accuracy Disparities in Commercial Gender Classification." *Proceedings of the 1st Conference on Fairness, Accountability and Transparency.* PMLR, volume 81, pp. 77-91.

13. Karl B. Christensen, Helene Feveile, Merete Labriola, and Thomas Lund, 2008. "The impact of psychosocial work environment factors on the risk of disability pension in Denmark." *European Journal of Public Health*, volume 18, number 3, pp. 235–237. doi: https://doi.org/10.1093/eurpub/ckm130

14. W.H.K. Chun, 2011. "Race and/as technology; or, how to do things to race." In: L. Nakamura and P. A. Chow-White (editors). *Race After the Internet*. New York: Routledge.

15. Dana S. Dunn and Erin E. Andrews, 2015. "Person-first and identity-first language: Developing psychologists' cultural competence using disability language." *The American Psychologist*, volume 70, number 3, pp. 255–264. doi: https://doi.org/10.1037/a0038636

16. Elizabeth Ellcessor, 2016. *Restricted access: Media, disability, and the politics of participation* (Vol. 6). New York: NYU Press.

17. Equal Employment Opportunity Commission, 2022. "U.S. EEOC and U.S. Department of Justice Warn against Disability Discrimination", at https://www.eeoc.gov/newsroom/us-eeoc-and-us-department-justice-warn-against-disability-discrimination , accessed 9 June 2022.

18. Michael Erard, 2017. "Why Sign-Language Gloves Don't Help Deaf People." *The Atlantic*. Retrieved from https://www.theatlantic.com/technology/archive/2017/11/why-sign-language-gloves-dont-help-deaf-people/545441/

19. Virginia Eubanks, 2018. *Automating Inequality: How High-Tech Tools Profile, Police, and Punish the Poor*. St. Martin's Press.

20. European Commission, 2020. "White paper on artificial intelligence: A European approach to excellence and trust." Com (2020) 65 Final.

21. Laura Forlano, 2017. "Data Rituals in Intimate Infrastructures : Crip Time and the Disabled Cyborg Body as an Epistemic Site of Feminist Science." *Catalyst: Feminism, Theory, Technoscience*, volume 3, number 2, pp. 1–28.

22. Miranda Fricker, 2007. *Epistemic injustice: Power and the ethics of knowing*. Oxford University Press.
31